\documentclass[lettersize,journal]{IEEEtran}
\usepackage{amsmath,amsfonts}
\usepackage{algorithm}
\usepackage{algpseudocode}
\usepackage{array}
\usepackage[caption=false,font=normalsize,labelfont=sf,textfont=sf]{subfig}
\usepackage{textcomp}
\usepackage{stfloats}
\usepackage{url}
\usepackage{verbatim}
\usepackage{graphicx}
\usepackage{tikz}
\usepackage{booktabs}
\usepackage{amssymb}
\usetikzlibrary{arrows.meta,positioning,calc,shapes.geometric, fit}
\usepackage{cite}
\newtheorem{definition}{Definition}

\newtheorem{remark}{Remark}

\newcommand{\MSE}{\operatorname{MSE}}

\begin{document}

\title{QueryMarket: Cost-Aware Online Active Learning in Data Markets}

% \author{Anonymous Author(s)}
\author{Xiwen~Huang and Pierre~Pinson,~\IEEEmembership{Fellow,~IEEE}%
\thanks{Xiwen Huang is with the Dyson School of Design Engineering, Imperial College London, London, SW7 2DB, U.K.
(e-mail: xiwen.huang23@imperial.ac.uk).}%
\thanks{Pierre Pinson is with the Dyson School of Design Engineering, Imperial College London, London, U.K. He has additional affiliations with Halfspace (part of Accenture), the Technical University of Denmark (DTU Management) and Aarhus University (CoRE) -- e-mail: p.pinson@imperial.ac.uk.}}

% \thanks{Manuscript received April 19, 2021; revised August 16, 2021.}}

% The paper headers
% \markboth{Journal of \LaTeX\ Class Files,~Vol.~14, No.~8, August~2021}%
% {Shell \MakeLowercase{\textit{et al.}}: A Sample Article Using IEEEtran.cls for IEEE Journals}
\markboth{}{}

% \IEEEpubid{0000--0000/00\$00.00~\copyright~2021 IEEE}
% Remember, if you use this you must call \IEEEpubidadjcol in the second
% column for its text to clear the IEEEpubid mark.

\maketitle

\begin{abstract}
Data acquisition is a major bottleneck for learning in real-time streams: analysts must decide \emph{on the fly} which labels to purchase while respecting a rolling budget. However, existing online active learning rarely unifies \emph{pricing}, \emph{information gain}, and \emph{rolling budget} constraints under concept drift. We introduce \textit{QueryMarket}, a market-inspired framework that queries each incoming data point based on its estimated utility to the model and its price. Within this framework, we propose OVBAL (online variance-based active learning), which integrates data pricing with information-driven selection by estimating each sample's marginal utility via a D-optimality criterion with exponential forgetting and executing cost-aware purchases under rolling budget constraints. OVBAL yields a simple, fully online decision rule that adapts to nonstationary streams and heterogeneous label costs. Experiments on synthetic data and a real-world solar power generation forecasting task show that OVBAL is particularly effective under seller-centric pricing and yields a more favorable long-run error--cost trade-off in the real-world task under both pricing schemes.

\end{abstract}

\begin{IEEEkeywords}
Active learning, data markets, mechanism design, regression, online learning
\end{IEEEkeywords}

\section{Introduction}

\subsection{Context and Motivation}

Machine learning models are increasingly being used in operational settings where data arrive sequentially and decisions must be made in real time \cite{shalev2012online,hoi2021online}. In supervised learning tasks, each observation consists of a feature vector and a corresponding ground-truth label. Although features are often readily available from sensors, digital platforms, or transactional systems, labels frequently require additional effort to obtain, such as laboratory measurements, expert annotations, audits, or external API queries. This asymmetry between easily observable inputs and costly ground-truth outcomes creates a fundamental bottleneck for learning systems, motivating selective label acquisition strategies as studied in active learning \cite{cacciarelli2024active,cohn1994active,settles2009survey}. This sequential acquisition setting resembles classical online selection, such as the secretary problem, where decisions must be made immediately without knowledge of future candidates. However, in the context of online learning, the objective is not to select a single optimal item, but to acquire a sequence of informative labels while managing a limited acquisition budget and adapting to the evolving state of the predictive model. 
Recent work on active learning markets for label purchase \cite{huang2026purchase} studied cost-effective label acquisition in a batch setting, where a fixed pool of unlabelled candidates is available and the analyst selects labels to purchase under explicit pricing constraints. Although the batch formulation establishes a principled connection between model improvement and economic cost, it relies on access to a static candidate set and does not model sequential arrivals. Moving to the online regime fundamentally changes the acquisition problem: label value becomes state-dependent as the model evolves, and each purchase decision incurs an opportunity cost by consuming budget that could be reserved for future, potentially more informative observations. Moreover, the marginal contribution of a label is not observable ex ante and must be estimated from the current model state. These considerations motivate an online market-aware acquisition rule that can price, select, and learn in real time from data streaming settings.

The online setting introduces additional challenges that are absent in batch acquisition mechanisms. Streaming environments often exhibit non-stationarity, where the underlying data-generating process evolves over time. Such changes may arise from seasonal effects, regime changes, sensor drift, or behavioral changes in the system being monitored. As a result, information obtained from older observations can gradually become outdated and less representative of the current environment \cite{shalev2012online,hoi2021online}. To adapt to such evolving conditions, online learning algorithms must discount old information while incorporating newly acquired observations efficiently. A widely adopted approach in recursive estimation and online regression is \emph{exponential forgetting}, which progressively reduces the weight on older samples and emphasizes recent data \cite{pinson2022regression, orabona2025modernintroductiononlinelearning}. Under a locally-stationary assumption, this mechanism can be interpreted as tracking a time-varying parameter vector, enabling the model to adapt to distributional changes while maintaining computational efficiency and numerical stability in sequential updates. This formulation is equivalent to assuming that model parameters evolve slowly over time, which allows the estimator to track gradual changes in the data-generating process. At the same time, label acquisition decisions must take into account sequential budget constraints. Most of the existing work on stream-based
active learning assumes a fixed query budget or limits the total
number of labels that can be requested \cite{fujii2016budgeted, cacciarelli2022online}. In contrast, in many practical data acquisition
scenarios, labels have heterogeneous prices, and budgets are spent
progressively over time. Each purchase, therefore, incurs an opportunity
cost by consuming resources that could otherwise be allocated to
future observations.

To address these issues, we propose \textit{QueryMarket}, where labels are owned by data sellers and can be purchased by a data analyst to improve predictive models. At each time step, a data seller provides a willingness-to-sell (WTS) for the current label, while a data analyst forms a willingness-to-pay (WTP) based on the estimated utility of the label for the predictive task. The analyst then decides whether to query the label by balancing informativeness and acquisition cost under a cumulative budget constraint. This leads to the central research question of this work: \textit{How should the analyst decide, in a streaming environment, whether to purchase the label when its utility is uncertain, and the acquisition budget must be managed sequentially over time?} Within QueryMarket, we develop an online variance-based active learning strategy (OVBAL) that estimates label utility from predictive uncertainty and performs sequential updates using a Newton-Raphson scheme with exponential forgetting. This design enables fast adaptation in non-stationary streams while maintaining explicit cost-efficiency in label acquisition. 

\subsection{Related Work}

\subsubsection{Active Learning (AL)} 

Although mainly studied for classification~\cite{balcan2010true, dasgupta2007general, hao2024composite, luth2023navigating, kontonis2024active}, AL is also studied in regression, in scenarios where labeling the output for the observations is costly~\cite{sabato2014active, sugiyama2006active}. AL falls into two broad categories: pool-based AL, where a fixed unlabeled data pool is available, and stream-based or online active learning (OAL), where data arrives sequentially \cite{settles.tr09}. Motivated by our setting in which data points arrive in a stream and labeling incurs a cost under a fixed budget, OAL provides the most natural framework. Although classical AL has a rich theoretical foundation~\cite{cohn1994active, cohn1994improving, balcan2010true,cacciarelli2024active, li2024survey}, recent efforts in OAL~\cite{goebel2025budgeted, saran2023streaming, fujii2016budgeted} remain limited in their treatment of real-world constraints. In particular, many OAL algorithms either assume labels are freely available or ignore economic considerations entirely. For example, threshold-based querying approaches~\cite{riquelme2017online} prioritize uncertainty but overlook the cost associated with each label. Building on the budget-aware querying rule for OAL in~\cite{goebel2025budgeted, cacciarelli2022stream}, we propose a more general and economically grounded framework, incorporating both seller-side willingness-to-sell and the analyst-side willingness-to-pay. This market-driven design enables our algorithm to adaptively balance informativeness with acquisition cost, enhancing efficiency under budget constraints.

\subsubsection{Data Market} 
As data becomes a tradeable asset, recent work in AI and machine learning has increasingly focused on \textit{data valuation} and \textit{data acquisition in markets}. Several approaches propose auction- or contract-based mechanisms to price data from strategic sellers~\cite{agarwal2019marketplace, ghosh2011selling, pandey2023strategic}. These methods typically assume batch settings, access to utility functions, or truthful revelations from sellers. In practice, however, label costs are heterogeneous, strategic, and often observable only at query time. Previous work on data valuation~\cite{amiri2023fundamentals} generally assumes offline access to data or task-agnostic settings, making them unsuitable for streaming environments. In contrast, our framework situates valuation within an on-line, cost-sensitive learning pipeline, and connects label acquisition decisions to expected utility through the lens of Information Value Theory~\cite{howard2007information}. 
% Another line of work explores incentive-compatible data purchasing~\cite{chawla2016incentivizing, zhao2021data}, sometimes incorporating privacy or fairness concerns~\cite{xu2021fair, zhao2023federated}. These methods often focus on mechanism design or reward allocation in federated or pooled environments, rather than online, individual-level selection. 
Closer to our setting, recent frameworks such as Data Shapley~\cite{ghorbani2019data} and the Data Valuation using Reinforcement Learning (DVRL) method~\cite{yoon2020data} estimate the marginal contribution of data points to task performance. However, they are generally retrospective and computationally intensive and require full retraining or simulation. In contrast, our approach introduces a lightweight \textit{real-time market-aware query strategy} tailored for streaming environments. By combining analyst-side WTP with seller-side WTS, we design an active selection rule that balances informativeness and cost efficiency without requiring utility ground truth or retraining. To our knowledge, this work provides one of the first frameworks that integrates active learning, market pricing, and information-theoretic utility estimation in an online setting.

\subsubsection{Online Learning} Our work also builds on the framework of online convex optimization (OCO)\cite{hoi2021online, shalev2012online, osogami2021second}, where the models are incrementally updated as new data arrive. Classical algorithms like online Gradient Descent (OGD)\cite{zinkevich2003online} and online Newton Step (ONS)\cite{hazan2007logarithmic} provide theoretical guaranties such as sublinear or logarithmic regret by leveraging first- or second-order information. The closest setting to ours is that of regression markets, where second-order methods such as Newton-Raphson updates with exponential forgetting are used to guide the purchase of features~\cite{pinson2022regression}. We extend this line of work to a different acquisition setting, QueryMarket, where the analyst purchases missing labels based on task-specific utility under budget and pricing constraints. Although regression markets focus on feature selection for model accuracy, our framework introduces an active learning market that combines utility estimation, cost awareness, and sequential query decisions. In contrast to existing approaches, our framework explicitly models the interaction between the value of data and the acquisition price in a sequential market environment.

\begin{figure*}[!ht]
\centering
\begin{tikzpicture}[
  font=\normalsize,
note/.style={font=\small, inner sep=1pt},
line width=0.5pt,
  >=Latex,
  node distance=7mm and 9mm,
  block/.style={
    draw, rounded corners=1.5pt,
    align=center,
    minimum height=9mm,
    minimum width=18mm,
    inner sep=2.2pt
  },
  smallblock/.style={
    draw, rounded corners=1.5pt,
    align=center,
    minimum height=7.5mm,
    minimum width=15mm,
    inner sep=2pt
  },
  db/.style={
    draw, cylinder, shape border rotate=90, aspect=0.25,
    align=center, minimum height=11mm, minimum width=13mm,
    inner sep=1pt
  },
  flow/.style={->, thick},
  aux/.style={->, thick, densely dashed}, % for payment/utility (grayscale-friendly)
  buyerframe/.style={draw, dashed, rounded corners=1.5pt, inner sep=3pt}
]

% --- Nodes (left -> right) ---
\node[smallblock] (stream) {Stream\\$\tilde{x}_t$};

\node[block, fill = blue!15, right=of stream] (model) {Forecast\\model};

\node[block, fill = blue!15, right=of model] (decision) {OVBAL decision\\ $u_t \ge \tau_{t}$,\ \ $\hat{\ell}_t \ge \eta_t/\phi$};

\node[db, above=8mm of decision, fill=gray!25] (seller) {\textbf{Seller}};

\node[smallblock, fill = blue!15, right=of decision] (update) {Update\\$\{\beta_t,H_t,\sigma^2_{y,t}\}$};

\node[block, right=of update] (task) {Downstream\\task(s)};

% --- Main information flow ---
\draw[flow] (stream) -- node[note, above] {$\tilde{x}_t$} (model);
\draw[flow] (model) --
  node[note, above] {$\hat{y}_t$}
  node[note, below=0.8mm] {$u_t, \ \hat{\ell}_t$}
(decision);
% Query branch: if query, get label and update
\draw[flow] (decision) -- node[note, above] {query} (update);
% label flow (seller -> decision), slight right shift
\draw[->, thick]
  ([xshift=1.3mm]seller.south) -- node[note, right] {$y_t$}
  ([xshift=1.3mm]decision.north);
% No-query branch: skip update (folding / forgetting)
\node[note, below=7mm of decision] (noq) {$\ H_t\!\leftarrow\!\lambda H_{t-1}$,\ $\beta_t\!\leftarrow\!\beta_{t-1}$};
\draw[flow] (decision.south) -- ++(0,-5mm) node[note, right] {} -- (noq.north);

% After update -> downstream
\draw[flow] (update) -- node[note, above] {$\hat{y}_{t+1}$} (task);

% --- Economic/utility flow (dashed, black) ---
% payment flow (decision -> seller), slight left shift, dashed
\draw[->, thick, densely dashed, red]
  ([xshift=-1.3mm]decision.north) -- node[left, font=\normalsize\bfseries, text=red] {payment $p_t$}
  ([xshift=-1.3mm]seller.south);
  % \draw[aux] (task.north) |- node[note, above] {utility / reward} (stream.north);

% --- Optional: compact decision rule annotation (kept small) ---
\node[note] (rule) at ($(decision.south)+(7mm,-3mm)$) {no query};
% --- timeline (aligned under the whole diagram) ---
\coordinate (timelineY) at ([yshift=-3mm]noq.south); % 这里调大/调小控制上下位置

\coordinate (timelineL) at ([xshift=-1mm]stream.west |- timelineY);
\coordinate (timelineR) at ([xshift= 1mm]task.east   |- timelineY);

\draw[->, line width=0.6pt] (timelineL) -- (timelineR)
  node[pos=1, below=2mm] {time};

% tick exactly ON the timeline under the decision block (t3)
\coordinate (tThree) at ($(timelineL)!(decision.south)!(timelineR)$);
\draw[line width=0.6pt] (tThree) ++(0,1.8mm) -- ++(0,-3.6mm)
  node[below=2mm] {$t_3$};

% optional: a few more ticks (remove if you want only t3)
\foreach \pos/\lab in {0.10/$t_1$,0.30/$t_2$,0.70/$t_4$,0.90/$t_T$}{
  \draw[line width=0.6pt] ($(timelineL)!\pos!(timelineR)$) ++(0,1.8mm) -- ++(0,-3.6mm)
    node[below=2mm] {\lab};
}
\end{tikzpicture}
\caption{
\textbf{QueryMarket architecture.} At time $t$, the analyst applies the OVBAL rule to decide whether to query a label from a seller. Queried labels are purchased at price $p_t$ (red dashed arrow) and used to update the forecasting model for downstream tasks. Blue blocks correspond to the buyer (analyst) and the grey block to the seller; black arrows indicate the data flow.}
\label{fig:market_intro}
\end{figure*}
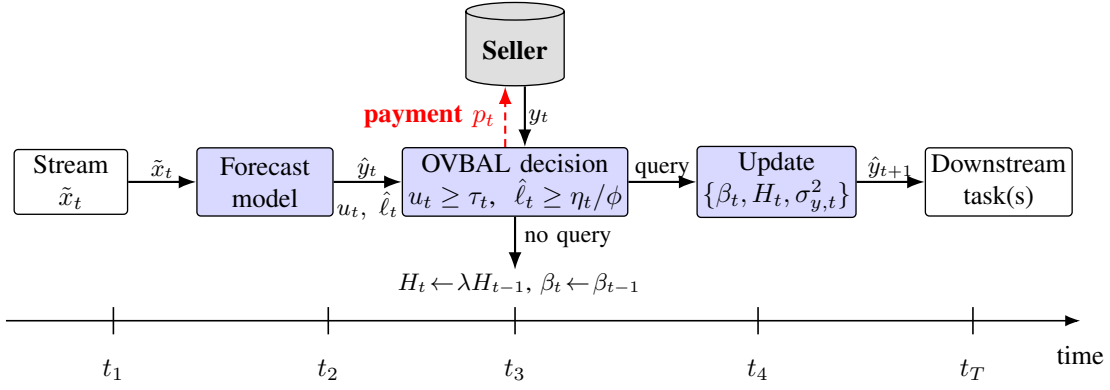

\subsection{Contributions}
Our contributions are threefold. First, we formulate streaming label acquisition as a market-aware online active learning problem with heterogeneous label prices, explicitly modelling seller-side willingness-to-sell and analyst-side willingness-to-pay under sequential spending constraints, thereby moving beyond the common unit-cost or fixed query-budget assumptions in active learning literature. Second, we develop OVBAL, a cost-aware querying strategy that couples uncertainty-based utility estimation with a budget-aware gating rule, and we integrate it with exponential forgetting to enable robust adaptation under non-stationarity. Third, we introduce an evaluation protocol for online active learning markets that compares acquisition strategies under equal-budget and prequential streaming metrics, and we use this protocol to benchmark buyer-centric and seller-centric pricing schemes on both synthetic and real-world datasets.

The remainder of the paper formalizes the QueryMarket setting and pricing interface, derives the OVBAL decision rule and its online update mechanism with forgetting, and evaluates the resulting market-aware acquisition pipeline on synthetic and real-world datasets. Across the experiments, the relative advantage of OVBAL depends on the pricing regime: it is particularly effective under seller-centric pricing, while in the real-world solar forecasting task, it delivers a more favorable long-run error--cost trade-off under both pricing schemes.
% Some recent 
\section{Problem Formulation}
\label{sec:problem}

We study \emph{cost-aware online label acquisition} in a streaming market, where labels are held by data sellers and must be purchased under a rolling budget.
At each discrete time step $t=1,\dots,T$, the analyst (buyer) first posts her WTP \(\phi\) and then observes the incoming covariates, needing to decide immediately whether to acquire the corresponding label; unpurchased labels expire and cannot be revisited. Let $z_t$ denote the raw covariates arriving at time $t$.
The analyst observes a (possibly transformed) feature vector $x_t\in\mathbb{R}^P$, while the ground-truth label $y_t\in\mathbb{R}$ is privately held by a seller.
The seller posts an ask price $\eta_t>0$ (WTS) for $y_t$ and the analyst will decide if she will buy it or not based on query strategy and decision role. This online interaction defines our \emph{QueryMarket}.

\subsection{Learning Model}
\label{subsec:model}

\paragraph{Linear regression setup}

We consider a streaming regression problem. At each time step \(t\), the analyst observes a feature vector \(x_t \in \mathbb{R}^p\). We use a linear predictor defined on the augmented feature vector
\begin{equation}
\tilde{x}_t = \begin{bmatrix}1 \\ x_t\end{bmatrix} \in \mathbb{R}^{p+1},
\end{equation}

where the leading 1 accounts for the intercept term. The prediction at time \(t\) is then given by
\begin{equation}
\hat{y}_t = f(x_t;\beta_{t-1}) = \beta_{t-1}^\top \tilde{x}_t,
\label{eq:pred_model}
\end{equation}
where \(\beta_{t-1} \in \mathbb{R}^{p+1}\) denotes the model parameter estimated from the labelled data available up to time \(t-1\). For each labelled observation, we measure the instantaneous prediction loss by
\begin{equation}
\ell\!\left(y_t,\hat{y}_t\right)
=
\ell\!\left(y_t,\beta_{t-1}^\top \tilde{x}_t\right).
\label{eq:inst_loss}
\end{equation}

In this work, we use the squared loss
\begin{equation}
\ell\!\left(y_t,\hat{y}_t\right) = \left(y_t-\hat{y}_t\right)^2.
\label{eq:sq_loss}
\end{equation}

The analyst is initially given a small labelled set \(D_L=\{(x_i,y_i)\}_{i=1}^{n_L},\)
which is used to initialize the predictor. Thereafter, observations arrive sequentially, and labels may only become available through the acquisition mechanism introduced below.
\paragraph{Nonlinearity via feature embeddings}
Although \eqref{eq:pred_model} is linear in $\tilde{x}_t$, it is \emph{not} restrictive with respect to the original covariates: $x_t$ can be any representation $x_t=\psi(z_t)$, where $\psi(\cdot)$ may be a nonlinear basis expansion (e.g., polynomials/splines), a kernel feature map (e.g., random Fourier features), or a learned embedding (e.g., a neural encoder).
Therefore, the overall mapping $z_t \mapsto \hat{y}_t$ can capture nonlinear relationships in the raw inputs while retaining an online-update-friendly form in the feature space.

\subsection{Market Interaction}
\label{subsec:market}

Let $\hat{\ell}_t$ denote the analyst's \emph{estimated utility} of acquiring the label at time $t$ (specified later via an information-gain criterion), and let $\phi>0$ denote the WTP per  unit of utility. The analyst submits a bid
\begin{equation}
\mathrm{bid}_t \;=\; \phi\,\hat{\ell}_t.
\label{eq:bid_def}
\end{equation}
A transaction is feasible only if the bid meets the ask, i.e.,
\begin{equation}
\mathrm{bid}_t \;\ge\; \eta_t.
\label{eq:feasible_trade}
\end{equation}
If a trade occurs, the label $y_t$ is revealed at transaction price $p_t$; otherwise, the opportunity expires.

We consider two pricing schemes:
\emph{seller pricing}, where $p_t=\eta_t$, and \emph{buyer pricing}, where $p_t=\mathrm{bid}_t$.
In both schemes, the feasibility condition \eqref{eq:feasible_trade} must hold.

% \subsection{Rolling Budget Mechanism}
% \label{subsec:budget}

% Unlike classical active learning with a fixed query budget, we adopt a rolling budget.
% Let $b_t^-$ be the available budget before the decision at time $t$, and $b_t^+$ the remaining budget after the decision.
% The budget evolves as
% \begin{equation}
% b_t^- = b_{t-1}^+ + \gamma,
% \qquad
% b_0^+ = b_0,
% \label{eq:budget_in}
% \end{equation}
% where $\gamma$ is the per-step income.
% A feasible purchase additionally requires $b_t^- \ge p_t$, in which case
% \begin{equation}
% b_t^+ = b_t^- - p_t;
% \label{eq:budget_out}
% \end{equation}
% otherwise (no purchase) $b_t^+=b_t^-$.
% We also track cumulative expenditure $c_t=\sum_{s\le t} p_s\,\mathbb{I}\{\text{trade at }s\}$.

\section{Proposed QueryMarket Framework} \label{sec:framework}

We propose \emph{Online Variance-Based Active Learning (OVBAL)}, a cost-aware query strategy for sequential label acquisition in streaming regression settings, where labels are selectively queried under budget constraints. OVBAL is designed to balance the informativeness of the queried data with the cost of acquisition, a setting motivated by practical constraints in real-time learning systems. Unlike traditional stream-based active learning methods that optimize A-optimality (minimizing average variance)~\cite{riquelme2017online}, OVBAL is built on the D-optimality principle, which selects queries to maximize information gain through determinant-based variance reduction. We operationalize this by estimating the \emph{Unscaled Predictive Variance} (UPV) as a proxy for sample utility and combine it with a market-aware pricing model involving the analyst's WTP and the seller's WTS. The result is a principled, cost-sensitive query rule that enables high-quality learning under limited budgets.

This method integrates three key components: (i) exponential forgetting to prioritize recent data; (ii) recursive Newton-Raphson updates for real-time model estimation; and (iii) a threshold-based utility estimator to drive query decisions. In addition, we adopt a rolling-budget market mechanism and an observer evaluation protocol, and select the forgetting factor via value-for-money cross-validation. The following subsections describe each component in detail.

\subsection{Exponential Forgetting}
\label{sec:method}

In dynamic environments where data distributions evolve over time, model adaptation must be both continuous and responsive. To accommodate non-stationarity in streaming data, we adopt an exponential forgetting factor \(\lambda \in [0,1)\), which emphasizes recent observations while discounting older ones. Therefore, the model parameters at time \( t \) are obtained by minimizing a time-discounted convex loss function over past observations:
\begin{align}
\hat{\beta}_{t} &= \arg\min_{\beta} L_{t}(\beta), \label{eq:beta_loss_function} \\
L_{t}(\beta) 
&= \frac{1}{n_\lambda} \sum_{t_i < t} \lambda^{t - t_i} \, \ell\left(y_{t_i}, \beta^\top \tilde{x}_{t_i}\right), \label{eq:loss_func}
\end{align}
where \( \ell(\cdot) \) is a convex loss function, \( \tilde{x}_{t} \) denotes the input vector augmented with a bias term, and \( n_\lambda = (1 - \lambda)^{-1} \) is the effective memory size. $L_{t}$ is a time-weighted estimator of the loss function $\ell$ evaluated over the time horizon up to $t$. In this work, we adopt the quadratic loss, i.e. \( \ell(y, \hat{y}) = (y - \hat{y})^2 \), which corresponds to the mean squared error (MSE).

\subsection{Recursive Updates}

Rather than recomputing the solution to~\eqref{eq:beta_loss_function} from scratch at each time step, we adopt a recursive Newton-style update. This builds on the framework of online convex optimization~\cite{hoi2021online}, where the analyst updates the parameters incrementally as new data arrive. Specifically, we incorporate ideas from the Newton-Raphson method ~\cite{pinson2022regression} and its extended online Newton Step (ONS) algorithm~\cite{hazan2007logarithmic, luo2016efficient}, which achieves logarithmic regret by using second-order curvature information in the update rule. We adapt these principles to a streaming regression setting with exponential forgetting.

Following~\cite{pinson2022regression}, given that \( \ell(\cdot) \) is a quadratic loss, we recursively update the model as follows:
\begin{subequations}
\begin{align}
\epsilon_t &= y_t - \hat{\beta}_{t-1}^\top \tilde{x}_t, \label{eq:model_update1}\\
H_t &= \lambda H_{t-1} + \tilde{x}_t \tilde{x}_t^\top, \label{eq:model_update2}\\
\hat{\beta}_t &= \hat{\beta}_{t-1} + H_t^{-1} \tilde{x}_t \epsilon_t, \label{eq:model_update3} \\
\sigma^2_{y,t} &= \lambda \sigma^2_{y,t-1} + (1 - \lambda)\epsilon_t^2. \label{eq:model_update4}
\end{align}
\end{subequations}

\subsection{Variance-Based Label Utility Estimation}
\paragraph{Uncertainty-Based Informativeness}  
Given a new input \( \tilde{x}_t \in \mathbb{R}^{p+1}\) at time \( t \), the UPV is computed as:
\begin{equation}
u_t = \tilde{x}_t^\top H_{t-1}^{-1} \tilde{x}_t,
\label{eq:upv}
\end{equation}
where $H_t$ is the Fisher information matrix maintained by recursive least squares updates, and
$u_t = \tilde x_t^\top H_{t-1}^{-1}\tilde x_t$ is the (unscaled) proxy of the predictive variance.
To adapt to distribution shift, we estimate a time-varying threshold $\tau_t$ using a rolling empirical quantile.
Specifically, let $\mathcal{U}_{t-1}=\{u_s\}_{s=\max(1,t-W)}^{t-1}$ denote the uncertainty history in a window of size $W$.
We set $\tau_t$ as the $(1-\alpha)$-quantile of $\mathcal{U}_{t-1}$:
\begin{equation}
\tau_t \;=\; q^{1-\alpha}\!\left(\mathcal{U}_{t-1}\right),
\label{eq:tau_rolling_quantile}
\end{equation}
so that the querying rate is controlled approximately by
\begin{equation}
\mathbb{P}(u_t \ge \tau_t)\approx \alpha.
\label{eq:query_rate_target}
\end{equation}
Only samples with uncertainty above this adaptive threshold are considered for labeling.

\begin{definition}[Cost-Aware Query Rule]
A label is acquired at time \( t \) if and only if:
\begin{align}
u_t &\geq \tau_t, \tag{Uncertainty Condition} \label{eq:cond1} \\
\hat{\ell}_t &\geq {\eta_t}/{\phi}. \tag{Cost-Utility Condition} \label{eq:cond2}
\end{align}
Here, \(\hat{\ell}_t\) denotes the estimated utility of a label, \( \phi \) denotes the analyst’s WTP per unit of utility, and \( \eta_t \) is the seller’s WTS.
\end{definition}

% --- Updated utility section: D-opt consistent (replace trace/A-opt) ---

In our QueryMarket, the analyst decides whether to purchase a label based on the expected utility of acquiring $y_t$.
Although the notion of utility can be task-specific (e.g. reducing downstream forecasting error or improving model stability), we adopt a principled, task-agnostic proxy: the expected reduction in model uncertainty induced by observation $x_t$.

\paragraph{D-optimal utility (expected log-volume reduction)}
In the linear-in-features model, the parameter covariance is proportional to the inverse information matrix.
Let $H_{t-1}$ denote the information matrix after processing data up to time $t-1$. Following \eqref{eq:model_update2}, we define \emph{estimated utility} as the incremental D-optimal log-gain from acquiring $(\tilde x_t,y_t)$, measured as the increase in log-determinant of the information matrix:
\begin{equation}
\begin{aligned}
\hat{\ell}_t \;=\;& \log\det(H_t) - \log\det(\lambda H_{t-1}) \\
=\;& \log\Bigl(1 + \tilde x_t^\top (\lambda H_{t-1})^{-1}\tilde x_t\Bigr) \\
=\;& \log\Bigl(1 + \tfrac{1}{\lambda}\tilde x_t^\top H_{t-1}^{-1}\tilde x_t\Bigr).
\end{aligned}
\label{eq:dopt_est_utility}
\end{equation}
where the second equality follows from the matrix determinant lemma.
\begin{remark}[Matrix determinant lemma]
For invertible $A$ and vector $u$, $\det(A+uu^\top)=\det(A)\big(1+u^\top A^{-1}u\big)$.
\end{remark}

\paragraph{True utility (realized D-opt gain)}
If the label is queried ($q_t=1$), we compute the \emph{true utility} as the realized reduction in model uncertainty measured by the D-optimal log-volume criterion.
We define the realized (global) utility as
\begin{equation}
\begin{aligned}
\ell_t \;=\;&
\log\det\bigl(\sigma_{y,t-1}^2 H_{t-1}^{-1}\bigr)
-\log\det\bigl(\sigma_{y,t}^2 H_t^{-1}\bigr) \\
=\;&
p\log\frac{\sigma_{y,t-1}^2}{\sigma_{y,t}^2}
+\Bigl(\log\det(H_t)-\log\det(H_{t-1})\Bigr).
\end{aligned}
\label{eq:dopt_true_utility}
\end{equation}
where $p=\dim(\tilde x_t)$ is the number of regression coefficients (including the intercept) and $\sigma_{y,t}^2$ is the online residual-variance estimate.

% --- Lemma placement suggestion (short, reviewer-proof) ---
% Put this as a Remark in the main text (1 line), and optionally a 3–4 line proof sketch in the appendix.
% \begin{remark}[Matrix determinant lemma]
% For invertible $A$ and vector $u$, $\det(A+uu^\top)=\det(A)\big(1+u^\top A^{-1}u\big)$.
% \end{remark}

\paragraph{Parameter uncertainty}
Under standard linear regression assumptions, the parameter covariance satisfies
\begin{equation}
\mathrm{Var}(\hat{\beta}_t)\;\approx\;\sigma^2_{y,t}\, H_t^{-1},
\label{eq:beta_var}
\end{equation}
where $\sigma^2_{y,t}$ is an online estimate of the residual variance.
In particular, the D-opt utility in~\eqref{eq:dopt_est_utility} depends only on $(\tilde x_t, H_{t-1}^{-1})$ at decision time (and does not require observing $y_t$),
making it time-consistent for streaming acquisition.

% ====== put this in Methodology, in this order ======
    \subsection{Online Acquisition Protocol}

\paragraph{Experimental protocol and data split}
We warm-start the model on an initial labeled set
\(D_L=\{(x_i,y_i)\}_{i=1}^{n_L}\).
The subsequent stream is divided into a validation segment
\(D_{\mathrm{cv}}=\{(x_t,\eta_t)\}_{t=1}^{T_{\mathrm{cv}}}\) to adjust the hyperparameters (e.g., \(\lambda\)),
and a hold-out evaluation stream
\(D_U=\{(x_t,\eta_t)\}_{t=T_{\mathrm{cv}}+1}^{T}\).
After selecting \(\lambda^\star\) on \(D_{\mathrm{cv}}\), we warm-start on \(D_L\cup D_{\mathrm{cv}}\) and run the online policy on \(D_U\). At each time \(t\), if the buyer purchases the label, she pays \(p_t\) and observes \(y_t\); otherwise \(y_t\) remains hidden to the analyst. To tune \(\lambda\), we define a criterion value-for-money (VfM) as the reduction in mean squared error (MSE) relative to a no-update baseline per unit acquisition cost:
\begin{equation}
\mathrm{VfM}(\lambda)
=
\frac{
\mathrm{MSE}_{\mathrm{no\mbox{-}update}}(D_{\mathrm{cv}})
-
\mathrm{MSE}_{\mathrm{alg}}(D_{\mathrm{cv}};\lambda)
}{
\sum_{t \in D_{\mathrm{cv}}} p_t(\lambda)
}.
\label{eq:vfm}
\end{equation}
where \(\mathrm{MSE}_{\mathrm{alg}}(D_\mathrm{cv};\lambda)\) denotes the running MSE obtained by the algorithm with forgetting factor \(\lambda\) on the validation stream \(D_\mathrm{cv}\), and \(\sum_{t \in D_{\mathrm{cv}}} p_t(\lambda)\) is the corresponding cumulative acquisition cost. For each algorithm, the forgetting parameter \(\lambda\) is
selected via validation by maximizing the value-for-money
(VfM) criterion.

\paragraph{Rolling budget}
We model a rolling budget that is replenished over time to reflect
sequential spending in streaming data acquisition.
Let \(b_t^-\) denote the available budget before the acquisition
decision at time \(t\), and \(b_t^+\) the remaining budget after the
decision, we have:
\(b_0^+=b_0\), \(b_t^- = b_{t-1}^+ + \gamma\).
A purchase is feasible only if \(b_t^- \ge p_t\); then \(b_t^+=b_t^- - p_t\).
If no purchase occurs, \(b_t^+=b_t^-\).
We also track cumulative spending \(c_t=\sum_{i=1}^t p_i\) for reporting.

\paragraph{Observer protocol and prequential MSE}
\label{subsec:observer}
We adopt an \emph{observer} protocol: at each step \(t\), the analyst makes a prediction using only the information available \emph{before} observing \(y_t\).
The realized outcome \(y_t\) is then revealed to the observer for evaluation, while it is used to update the analyst \emph{only if} the label is purchased.
Let \(\hat{y}_t=f(x_t;\hat{\beta}_{t-1})\) and define the prequential error \(e_t = y_t-\hat{y}_t\).
We report the running  MSE,
whose terminal value is equal to the overall MSE on a segment \(\mathcal{S}\):
\begin{equation}
  \MSE(\mathcal{S}) \;=\; \frac{1}{|\mathcal{S}|}\sum_{t\in \mathcal{S}} e_t^2
  \;=\; \frac{1}{|\mathcal{S}|}\sum_{t\in \mathcal{S}} (y_t-\hat{y}_t)^2 .
  \label{eq:preq_mse}
\end{equation}
Lower values indicate better predictive accuracy, and the running curve illustrates stabilization speed and sensitivity to non-stationarity.

\paragraph{Query execution and pricing}
When the buyer decides to query, the transaction price \(p_t\) follows one of two pricing rules:
\begin{equation}
p_t =
\begin{cases}
\phi \hat{\ell}_t & \text{(buyer-centric pricing)}\\
\eta_t & \text{(seller-centric pricing)}.
\end{cases}
\label{eq:pricing}
\end{equation}
Label purchase occurs only when it is budget-feasible, i.e. \(b_t^- \ge p_t\);
under buyer-centric pricing, we additionally require the seller to accept the bid \(\phi \hat{\ell}_t \ge \eta_t\).
If a purchase occurs, the analyst updates \(\hat{\beta}_t\), \(H_t\), and \(\sigma^2_{y,t}\) via \eqref{eq:model_update1}--\eqref{eq:model_update4};
otherwise it performs only the forgetting step (no label update).
\begin{algorithm}[t]
\caption{Online Variance-Based Active Learning (OVBAL)}
\label{alg:OVBAL}
\begin{algorithmic}[1]
\State \textbf{Notation:} $[\cdot]\in\{0,1\}$; $\tilde{x}_t \triangleq [1;\,x_t]$
\State \textbf{Input:} labeled $D_L$, stream $\{(x_t,\eta_t)\}_{t=1}^{|D_U|}$,  labels $\{y_t\}$
\State \textbf{Params:} forgetting factor $\lambda$, threshold $\alpha$, WTP $\phi$, initial budget $b_0$, per-step income $\gamma$, window $W$
\State \textbf{Init:} $H_0,\hat{\beta}_0,\sigma^2_{y,0}$; $b_0^+\gets b_0$; $c\gets 0$; $\mathcal{U}\gets \mathcal{U}(D_L)$; $\tau_0$ by \eqref{eq:tau_rolling_quantile}
\State \textbf{Output:} $\{\hat{\beta}_t\}$, $\{\hat{y}_t\}$, $\{p_t\}$, $c$, $b^+_{|D_U|}$

\For{$t=1$ \textbf{to} $|D_U|$}
  \State $b_t^- \gets b_{t-1}^+ + \gamma$
  \State $\hat{y}_t \gets \hat{\beta}_{t-1}^\top \tilde{x}_t$, \quad $u_t \gets \tilde{x}_t^\top H_{t-1}^{-1}\tilde{x}_t$
  \State $\tau_{t-1} \gets \tau(\mathcal{U})$ via \eqref{eq:tau_rolling_quantile}
  \State $\hat{\ell}_t$ by \eqref{eq:dopt_est_utility}; \ $p_t$ by \eqref{eq:pricing}
  \State $q_t \gets [u_t \ge \tau_{t}]\,[\phi\hat{\ell}_t \ge \eta_t]\,[b_t^- \ge p_t]$

  \If{$q_t=1$}
    \State purchase $y_t$; \ $c\gets c+p_t$; \ $b_t^+ \gets b_t^- - p_t$
    \State Update $(\epsilon_t,H_t,\hat{\beta}_t,\sigma^2_{y,t})$ via \eqref{eq:model_update1}--\eqref{eq:model_update4}
    \State compute $\ell_t$ via \eqref{eq:dopt_true_utility}
  \Else
    \State $p_t\gets 0$, $\ell_t\gets 0$, $b_t^+\gets b_t^-$
    \State $H_t\gets \lambda H_{t-1}$,\ $\hat{\beta}_t\gets \hat{\beta}_{t-1}$,\ $\sigma^2_{y,t}\gets \sigma^2_{y,t-1}$
  \EndIf

  \State $\mathcal{U}\gets \textsc{Tail}_W(\mathcal{U}\cup\{u_t\}),\ \tau_t \gets \tau(\mathcal{U})$ via \eqref{eq:tau_rolling_quantile}
\EndFor
\end{algorithmic}
\end{algorithm}

We summarize the complete OVBAL algorithm in Algorithm~\ref{alg:OVBAL}. To evaluate the effectiveness of our OVBAL strategy, we compare it with two baselines: a Greedy strategy and a Random Sampling (RS) strategy, described in Algorithms~\ref{alg:Greedy} and~\ref{alg:RS}, respectively. In \textsc{Greedy}, the buyer purchases every arriving label whenever it is budget-feasible, i.e., \(b_t^- \ge p_t\), without incorporating any informativeness or utility-based selection criterion.
In \textsc{RS}, the buyer purchases a label with fixed probability \(p\), again only when \(b_t^- \ge p_t\), providing a constant-rate acquisition benchmark under the same rolling-budget constraint. These baselines isolate the contribution of OVBAL's utility-driven selection beyond budget feasibility alone.

\begin{algorithm}[t]
\caption{Greedy}
\label{alg:Greedy}
\begin{algorithmic}[1]
\State \textbf{Input:} labeled $D_L$, stream $\{(x_t,\eta_t)\}_{t=1}^{|D_U|}$, labels $\{y_t\}$
\State \textbf{Parameters:} $\lambda$, $\phi$, $b_0$, $\gamma$
\State \textbf{Init:} $H_0,\hat{\beta}_0,\sigma^2_{y,0}$; $b_0^+\gets b_0$; $c\gets 0$
\State \textbf{Output:} $\{\hat{\beta}_t\}$, $\{\hat{y}_t\}$, $\{p_t\}$, $c$, $b^+_{|D_U|}$

\For{$t=1$ \textbf{to} $|D_U|$}
  \State $b_t^- \gets b_{t-1}^+ + \gamma$
  \State Predict $\hat{y}_t$ using $\hat{\beta}_{t-1}$
  \State Compute price $p_t$ by \eqref{eq:pricing}
  \If{$b_t^- \ge p_t$}
    \State Purchase $y_t$, $c \gets c + p_t$, \quad $b_t^+ \gets b_t^- - p_t$
    \State Update model via \eqref{eq:model_update1}--\eqref{eq:model_update4}
  \Else
    \State $p_t \gets 0$, \quad $b_t^+ \gets b_t^-$
    \State $H_t \gets \lambda H_{t-1},\ \hat{\beta}_t \gets \hat{\beta}_{t-1},\ \sigma^2_{y,t} \gets \sigma^2_{y,t-1}$
  \EndIf
\EndFor
\end{algorithmic}
\end{algorithm}

% ---- Random Sampling (RS) with rolling budget ----
\begin{algorithm}[t]
\caption{Random Sampling (RS)}
\label{alg:RS}
\begin{algorithmic}[1]
\State \textbf{Input:} labeled $D_L$, stream $\{(x_t,\eta_t)\}_{t=1}^{|D_U|}$, labels $\{y_t\}$
\State \textbf{Parameters:} $\lambda$,  $\phi$, $b_0$,  $\gamma$, query probability $p$
\State \textbf{Init:} $H_0,\hat{\beta}_0,\sigma^2_{y,0}$; $b_0^+\gets b_0$;  $c\gets 0$
\State \textbf{Output:} $\{\hat{\beta}_t\}$, $\{\hat{y}_t\}$, $\{p_t\}$, $c$, $b^+_{|D_U|}$

\For{$t=1$ \textbf{to} $|D_U|$}
  \State $b_t^- \gets b_{t-1}^+ + \gamma$
  \State Sample query indicator $q_t \sim \mathrm{Bernoulli}(p)$
  \State Predict $\hat{y}_t$ using $\hat{\beta}_{t-1}$; compute price $p_t$ by \eqref{eq:pricing}
  \If{$q_t=1$ \textbf{and} $b_t^- \ge p_t$}
    \State Purchase $y_t$; $c \gets c + p_t$, \quad $b_t^+ \gets b_t^- - p_t$
    \State update model via \eqref{eq:model_update1}--\eqref{eq:model_update4}
  \Else
    \State $p_t \gets 0$, \quad $b_t^+ \gets b_t^-$
    \State $H_t \gets \lambda H_{t-1},\ \hat{\beta}_t \gets \hat{\beta}_{t-1},\ \sigma^2_{y,t} \gets \sigma^2_{y,t-1}$
  \EndIf
\EndFor
\end{algorithmic}
\end{algorithm}

\section{Numerical results}
In this section, we first use a synthetic setting to illustrate the behavior of OVBAL and the benchmark methods under controlled pricing regimes, and then evaluate the methods on a real-world solar power forecasting task. While the synthetic study is useful for interpreting the acquisition mechanisms, the real-data experiment provides the main evidence of practical performance.

\subsection{Synthetic Data Setting}
We use this synthetic experiment primarily to illustrate how the acquisition mechanisms behave under controlled buyer- and seller-pricing regimes before turning to the real-world forecasting task. We consider a time-varying linear regression model. At each time step $t \in \{1,\dots,T\}$, we observe a feature vector
$\mathbf{x}_t = (x_{t,1},x_{t,2},x_{t,3})^\top \in \mathbb{R}^3$ with i.i.d.\ components $x_{t,i} \sim \mathcal{N}(0,1)$.
The response is generated as
\begin{equation}
    y_t = \beta_{0,t} + \beta_{1,t} x_{t,1} + \beta_{2,t} x_{t,2} + \beta_{3,t} x_{t,3} + \epsilon_t,
\end{equation}
where $\epsilon_t \sim \mathcal{N}(0,\sigma_\epsilon^2)$ with $\sigma_\epsilon = 0.3$, and $T=20000$.
The time-varying coefficients follow a smooth drift:
\begin{align}
    \beta_{0,t} &= 1 + 2\times 10^{-4}\, t,\\
    \beta_{1,t} &= \sin(0.005\, t),\\
    \beta_{2,t} &= 0.5 + 10^{-3}\, t,\\
    \beta_{3,t} &= 3\times 10^{-4}\, t.
\end{align}

At each $t$, the buyer posts a bid proportional to her WTP, implemented as
$b_t = \phi \,\hat{\ell}_t$ with $\phi=0.16$, where $\hat{\ell}_t$ is calculated from \eqref{eq:dopt_est_utility}.
The seller's ask price $\eta_t$ is generated i.i.d.\ as a scaled log-normal random variable: $\eta_t = c\,\tilde{\eta}_t$, where $\tilde{\eta}_t \sim \mathrm{LogNormal}(\mu=-2,\sigma=0.30)$ and $c=16$. The agent starts with budget $b_0=1.6 \times 10^{-2}$ and receives a fixed income $\gamma=6.4\times 10^{-3}$ at each step. For OVBAL, the selection threshold $\tau_t$ is computed as the rolling $85$th percentile of the most recent 200 historical uncertainty scores $\{u_{t'}\}$. Under stable conditions, this yields an approximate candidate rate comparable to querying the top $15\%$ of high-uncertainty observations. We divide the stream chronologically into labelled data $D_L$, validation data $D_{\mathrm{cv}}$, and test data $D_U$, using the earliest $5\%$ of observations for warm-starting, the subsequent $10\%$ for hyperparameter tuning, and the remaining $85\%$ for evaluation. The forgetting factor $\lambda$ is tuned in $D_{\mathrm{cv}}$ on a grid
$\{0.990,0.991,\ldots,0.999\}$ by maximizing value-for-money (VfM), and then the selected $\lambda$ is used for evaluation in $D_U$.
\begin{table*}[t]
\centering
\caption{Overall cost and average cost per acquired data point (£/data) on the synthetic dataset under BC (buyer) and SC (seller) pricing.}
\label{tab:synth_cost_bc_sc}
\begin{tabular}{lrrrrrr}
\toprule
& \multicolumn{3}{c}{BC (buyer)} & \multicolumn{3}{c}{SC (seller)} \\
\cmidrule(lr){2-4} \cmidrule(lr){5-7}
Method 
& Overall cost (£) & Data & £/data
& Overall cost (£) & Data & £/data \\
\midrule
OVBAL  & 102   & 79    & 1.29                & 89.94 & 79 & 1.14 \\
Greedy & 10.19 & 17000 & $6.00\times10^{-4}$ & 108.5 & 97 & 1.12 \\
RS     & 9.01  & 2659  & $3.39\times10^{-3}$ & 108.5 & 80 & 1.36 \\
\bottomrule
\end{tabular}
\end{table*}

\begin{figure*}[t]
\centering
\subfloat[Buyer pricing]{%
  \includegraphics[width=0.48\textwidth]{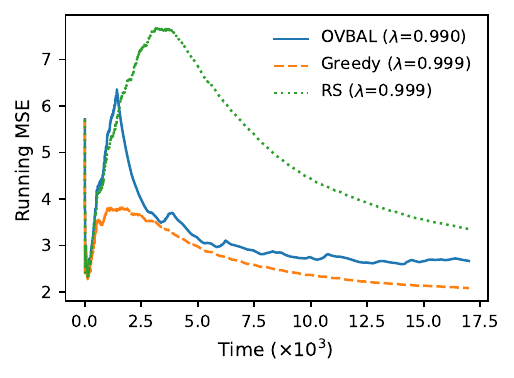}%
  \label{fig:toy_mse_buyer}}
\hfill
\subfloat[Seller pricing]{%
  \includegraphics[width=0.48\textwidth]{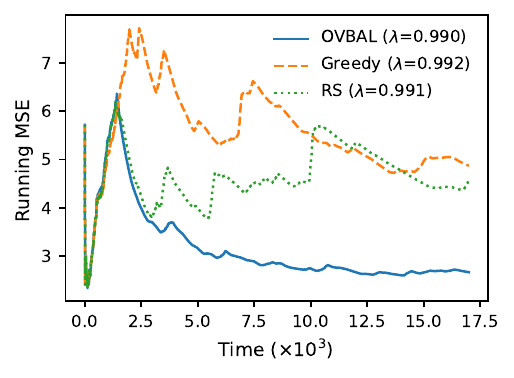}%
  \label{fig:toy_mse_seller}}
\caption{Synthetic data: running prequential MSE versus time step under buyer (BC) and seller (SC) pricing.}
\label{fig:toy1}
\end{figure*}

\begin{figure*}[t]
\centering
\subfloat[Buyer pricing]{%
  \includegraphics[width=0.48\textwidth]{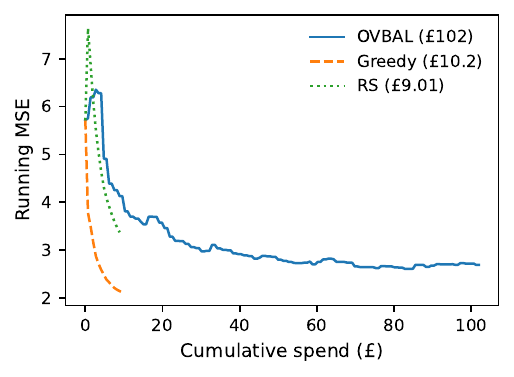}}
\hfill
\subfloat[Seller pricing]{%
  \includegraphics[width=0.48\textwidth]{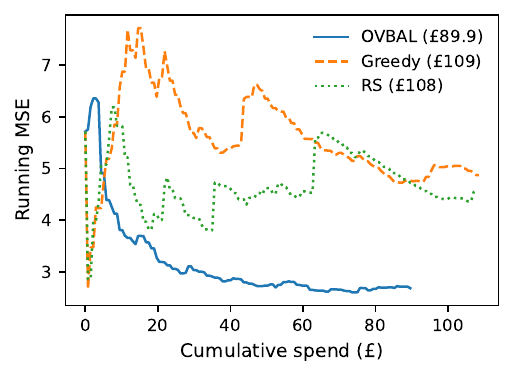}%
  \label{fig:budget-seller}}
\caption{Synthetic data: MSE versus cumulative spend (equal-budget comparison) under buyer (BC) and seller (SC) pricing.}
\label{fig:budget}
\end{figure*}
Figure~\ref{fig:toy1} reports the running prequential MSE over time on the synthetic data set under buyer-centric (BC) and seller-centric (SC) pricing. The results differ markedly across the two pricing regimes. Under BC pricing (Fig.~\ref{fig:toy_mse_buyer}), the greedy strategy achieves the lowest running MSE after the initial transient phase, while OVBAL performs better than random sampling but does not outperform Greedy. This behavior is consistent with the BC mechanism implemented here: since the paid price equals the learner's bid, Greedy can acquire a very large number of low-bid labels while incurring only a low total cost. As a result, in this controlled setting, broad low-cost acquisition proves more effective than selective high-utility acquisition. In contrast, OVBAL is much more selective and purchases only a small number of high-utility but high-bid labels, resulting in a substantially higher average cost per acquired label.

Under SC pricing (Fig.~\ref{fig:toy_mse_seller}), OVBAL clearly achieves the lowest running MSE throughout most of the stream. In this regime, prices are exogenously determined by sellers, so acquisitions require a stronger utility--cost trade-off. As a result, indiscriminate querying becomes inefficient, and OVBAL's D-optimality-driven selection yields a clear advantage over both Greedy and RS.

Table~\ref{tab:synth_cost_bc_sc} further illustrates these regime-dependent acquisition patterns. Under BC pricing, OVBAL acquires only 79 labels but incurs the highest total cost (£102), whereas Greedy purchases 17,000 labels at a much lower total cost (£10.19) and attains the best predictive accuracy. Thus, in this synthetic BC setting, OVBAL is more selective but not more cost-effective than Greedy. Under SC pricing, however, OVBAL achieves both the lowest overall acquisition cost (£89.94) and the lowest prediction error, while Greedy and RS incur higher total spending (£108.5) and worse forecasting performance.

Figure~\ref{fig:budget} compares the methods at equal budget levels by plotting running MSE against cumulative spending. The equal-budget comparison again highlights the difference between the two pricing regimes. Under BC pricing, OVBAL does not provide the best error--cost trade-off: over the common expenditure range, Greedy achieves lower running MSE at substantially lower total cost. Under SC pricing, by contrast, OVBAL consistently delivers the most favorable error--cost trade-off, achieving lower prediction error than Greedy and RS over most of the comparable spending range. Overall, this synthetic experiment shows that OVBAL is particularly effective when acquisition prices are externally imposed and heterogeneous, whereas under the present BC setup, Greedy benefits from being able to purchase many labels at very low endogenous bid prices. This further highlights that the relative advantage of acquisition strategies depends on the pricing mechanism and data environment, motivating the real-world evaluation in the next subsection.

\subsection{Solar Power Forecasting}

We evaluated OVBAL on a real-world solar power forecasting task using the UNISOLAR data set \cite{wimalaratne2022unisolar}, which contains photovoltaic (PV) generation measurements from 42 solar sites distributed across five campuses, recorded at a 15-minute resolution between January 2020 and April 2022. Experiments were conducted on \textbf{Site 10 (Campus 3)} of the UNISOLAR data set, selected for its relatively complete meteorological records and limited missingness. The site provides synchronized PV generation and weather measurements, including apparent temperature, air temperature, dew point temperature, relative humidity, wind speed, and wind direction.

The forecasting task is one-step-ahead prediction, i.e., predicting PV generation 15 minutes ahead. Measurements are first aligned to a regular 15-minute grid. Nighttime generation is fixed to zero based on astronomical sunrise--sunset calculations, and only daytime samples are retained for modelling. The input features consist of contemporaneous meteorological variables, sine--cosine encoding of wind direction, cyclic temporal features for hour-of-day, day-of-year, and month-of-year, and three autoregressive PV lags corresponding to the previous 15, 30, and 60 minutes, i.e., $y_{t-1}$, $y_{t-2}$, and $y_{t-4}$. All input features are standardized using the mean and standard deviation computed from the initial labelled warm-up subset only.

We fix the market parameters to $\phi=0.38$, initial budget $b_0=10^{-3}$, and per-step budget replenishment $\gamma_{\text{income}}=2\times 10^{-3}$. For each time step, the ask price is sampled i.i.d. from a log-normal distribution, $\eta_t \sim \operatorname{LogNormal}(\mu_\eta,\sigma_\eta^2)$, with $\mu_\eta=-2$ and $\sigma_\eta=0.30$.

Data are split chronologically to preserve temporal structure: the earliest $5\%$ of observations form the initial labelled set used to warm-start the model, and the subsequent $10\%$ are used for hyperparameter tuning. For buyer-centric (BC) pricing, the optimal forgetting factors are $\lambda=0.995$ (OVBAL), $\lambda=0.998$ (Greedy), and $\lambda=0.999$ (RS), while for seller-centric (SC) pricing they are $\lambda=0.999$ (OVBAL), $\lambda=0.996$ (Greedy), and $\lambda=0.998$ (RS). The remaining observations are then processed sequentially as an unlabelled stream to simulate online operation.

We compare OVBAL against (i) a greedy baseline that queries whenever budget remains available (\emph{Greedy}) and (ii) a random sampling baseline that queries each incoming data point with probability $0.15$ (\emph{RS}(15\%)). For OVBAL, we set $\alpha=0.15$ and use a rolling window of 200 observations to compute the threshold $\tau_t$ in \eqref{eq:tau_rolling_quantile}. This yields an approximate candidate rate comparable to RS(15\%) under stable conditions. Since the estimated D-optimal utility $\hat{\ell}_t$ is monotone in $u_t$ in \eqref{eq:dopt_est_utility}, this thresholding rule is consistent with the utility-based interpretation of OVBAL. All methods share the same budget replenishment mechanism and are evaluated under the same pricing setup with each regime.

Table~\ref{tab:solar_cost} reports the overall acquisition cost and the
average cost per acquired label under buyer-centric (BC) and
seller-centric (SC) pricing on the solar forecasting task. Under both
pricing schemes, OVBAL acquires far fewer labels than the Greedy
baseline. In BC pricing, OVBAL purchases only 47 labels, compared with
662 for Greedy and 802 for RS, while incurring a similar overall
expenditure. In SC pricing, OVBAL acquires 458 labels, substantially
fewer than Greedy (576) and nearly the same as RS (456). These results
indicate that OVBAL achieves stronger forecasting accuracy without
requiring more total spending and without relying on substantially more
label acquisitions than the baselines.

\begin{figure*}[t]
\centering

\subfloat[Buyer pricing]{%
    \includegraphics[width=0.45\textwidth]{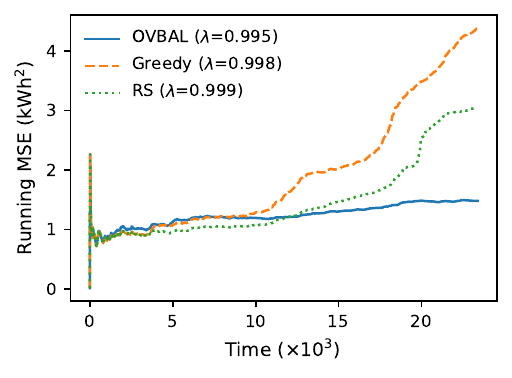}%
    \label{fig:solar_running_buyer}
}
\hfill
\subfloat[Seller pricing]{%
    \includegraphics[width=0.45\textwidth]{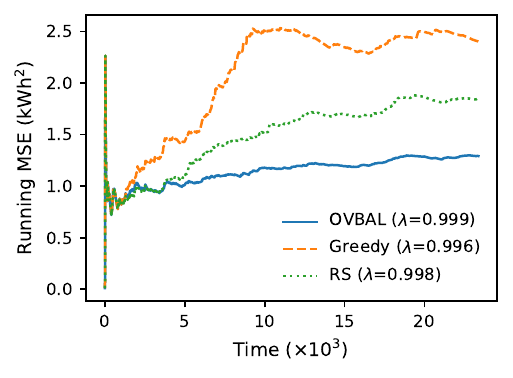}%
    \label{fig:solar_running_seller}
}

\caption{Solar forecasting: running MSE over time under buyer (BC) and seller (SC) pricing.}
\label{fig:solar_running}
\end{figure*}

\begin{figure*}[t]
\centering

\subfloat[Buyer pricing]{%
    \includegraphics[width=0.45\textwidth]{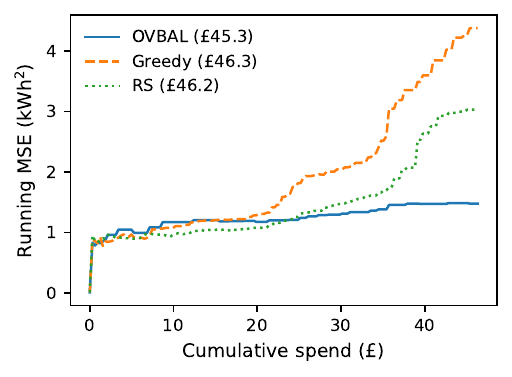}%
    \label{fig:solar_budget_buyer}
}
\hfill
\subfloat[Seller pricing]{%
    \includegraphics[width=0.45\textwidth]{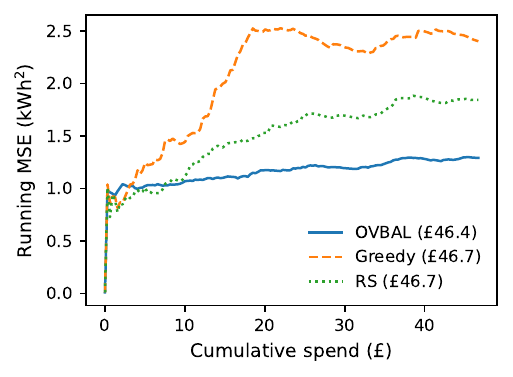}%
    \label{fig:solar_budget_seller}
}
\caption{Solar forecasting: equal-budget MSE versus cumulative spend under buyer (BC) and seller (SC) pricing. Numbers in parentheses denote the final cumulative spend.}
\label{fig:solar_budget}
\end{figure*}

Fig.~\ref{fig:solar_running} shows the running MSE over time under BC
and SC pricing. Across both pricing schemes, OVBAL generally maintains the lowest and most stable prediction error over the streaming horizon. Greedy
achieves a relatively fast initial reduction in error, but its
performance deteriorates later, especially under BC pricing, suggesting
that aggressive early acquisitions may lead to inefficient budget use in
the long run. This pattern is
consistent with the buyer-pricing mechanism, under which missed
informative labels may allow uncertainty to accumulate along
poorly learned directions. By prioritizing samples with higher
uncertainty-reduction value, OVBAL mitigates this effect and
maintains better long-run performance. RS performs better than Greedy in some periods but remains
consistently worse than OVBAL overall. To isolate acquisition efficiency from total spending, Fig.~\ref{fig:solar_budget} plots the cumulative running MSE against cumulative spend. Since the vertical axis is the average prediction error accumulated up to the time when a given spending level is reached, the comparison is path-dependent. OVBAL is therefore not always best in the very low-spend regime, where Greedy or RS may achieve slightly lower cumulative error through more aggressive early acquisitions. However, as the stream progresses, OVBAL delivers a more favorable long-run error--cost trade-off and attains the lowest final running MSE.

% Under buyer pricing, the learner’s bid is proportional to the estimated D-optimal utility gain, i.e., $\mathrm{bid}_t=\phi\,\wide\hat{\ell}_t$. By \eqref{eq:dopt_est_utility}, $\wide\hat{\ell}_t=\log\!\bigl(1+\tfrac{1}{\lambda}\tilde x_t^\top H_{t-1}^{-1}\tilde x_t\bigr)$ increases with directional uncertainty $\tilde x_t^\top H_{t-1}^{-1}\tilde x_t$. When informative labels are missed, the model cannot be corrected and $H_{t-1}^{-1}$ may inflate along poorly learned directions, increasing $\wide\hat{\ell}_t$ (and hence $\mathrm{bid}_t$) until bids exceed the rolling budget; this may block further acquisitions and induce a self-reinforcing deadlock with bursts of large errors. In contrast, OVBAL targets samples that most reduce uncertainty per acquired label, enabling more strategic purchases and preserving budget buffer, which empirically mitigates buyer-pricing deadlock.
\begin{table*}[t]
\centering
\caption{Overall cost and average cost per acquired data point (£/data) on the real-world solar forecasting dataset under BC and SC pricing.}
\label{tab:solar_cost}
\begin{tabular}{lrrrrrr}
\toprule
& \multicolumn{3}{c}{BC (buyer)} & \multicolumn{3}{c}{SC (seller)} \\
\cmidrule(lr){2-4} \cmidrule(lr){5-7}
Method 
& Overall cost (£) & Data & £/data
& Overall cost (£) & Data & £/data \\
\midrule
OVBAL  & 45.33 & 47  & 0.964 & 46.41 & 458 & 0.101 \\
Greedy & 46.31 & 662 & 0.070 & 46.71 & 576 & 0.081 \\
RS     & 46.16 & 802 & 0.058 & 46.73 & 456 & 0.102 \\
\bottomrule
\end{tabular}
\end{table*}
% ==============================
% Running MSE
% ==============================

% ==============================
% Equal-budget MSE
% ==============================

% \begin{figure*}[t]
% \centering

% \subfloat[Buyer pricing]{%
%     \includegraphics[width=0.45\textwidth]{figs_overleaf/y_vs_yhat_full_OVBAL_buyer.pdf}%
%     \label{fig:ovbal-buyer}
% }
% \hfill
% \subfloat[Seller pricing]{%
%     \includegraphics[width=0.45\textwidth]{figs_overleaf/y_vs_yhat_full_OVBAL_seller.pdf}
%     \label{fig:ovbal-seller}
% }
% \caption{Solar forecasting: predict and real difference (OVBAL)}
% \label{fig:solar_budget}
% \end{figure*}

% \begin{figure*}[t]
% \centering

% \subfloat[Buyer pricing]{%
%     \includegraphics[width=0.45\textwidth]{figs_overleaf/y_vs_yhat_full_Greedy_buyer.pdf}%
%     \label{fig:greedy-buyer}
% }
% \hfill
% \subfloat[Seller pricing]{%
%     \includegraphics[width=0.45\textwidth]{figs_overleaf/y_vs_yhat_full_Greedy_seller.pdf}
%     \label{fig:greedy-seller}
% }
% \caption{Solar forecasting: predict and real difference (Greedy)}
% \label{fig:solar_budget}
% \end{figure*}

% \begin{figure*}[t]
% \centering

% \subfloat[Buyer pricing]{%
%     \includegraphics[width=0.45\textwidth]{figs_overleaf/y_vs_yhat_full_RS_buyer.pdf}%
%     \label{fig:rs-buyer}
% }
% \hfill
% \subfloat[Seller pricing]{%
%     \includegraphics[width=0.45\textwidth]{figs_overleaf/y_vs_yhat_full_RS_seller.pdf}
%     \label{fig:rs-seller}
% }
% \caption{Solar forecasting: predict and real difference (RS)}
% \label{fig:solar_budget}
% \end{figure*}

\section{Conclusion}

In this paper, we proposed \textit{QueryMarket}, an online market-aware framework for cost-effective label acquisition in streaming environments. Building on our earlier batch active learning market formulation, QueryMarket extends the label-purchasing problem to the online setting, where acquisition decisions must be made sequentially under uncertainty and budget constraints. The framework models the interaction between the data analyst and data sellers through willingness-to-pay and willingness-to-sell, thereby incorporating heterogeneous label prices into the active learning process. Within this framework, we developed OVBAL, a cost-aware online acquisition strategy that combines utility-based querying with exponential forgetting to adapt to non-stationary data streams. This design enables the analyst to account for both the informativeness and the acquisition cost of candidate labels while maintaining efficient online model updates. Experimental results on synthetic and real-world datasets demonstrated the effectiveness of the proposed approach. In particular, the synthetic study showed that the benefit of OVBAL depends on the pricing regime, with the clearest advantage arising under seller-centric pricing, while the solar power forecasting case study showed that OVBAL delivers a more favorable long-run error--cost trade-off under both buyer-centric and seller-centric pricing schemes.
\bibliographystyle{IEEEtran}
\bibliography{myref}

@inproceedings{riquelme2017online,
  title={Online active linear regression via thresholding},
  author={Riquelme, Carlos and Johari, Ramesh and Zhang, Baosen},
  booktitle={Proceedings of the AAAI Conference on Artificial Intelligence},
  volume={31},
  number={1},
  year={2017}
}

@article{cacciarelli2022online,
  title={Online active learning for soft sensor development using semi-supervised autoencoders},
  author={Cacciarelli, Davide and Kulahci, Murat and Tyssedal, John},
  journal={arXiv preprint arXiv:2212.13067},
  year={2022}
}

@inproceedings{amiri2023fundamentals,
  title={Fundamentals of task-agnostic data valuation},
  author={Amiri, Mohammad Mohammadi and Berdoz, Frederic and Raskar, Ramesh},
  booktitle={Proceedings of the AAAI Conference on Artificial Intelligence},
  volume={37},
  number={8},
  pages={9226--9234},
  year={2023}
}

@article{sugiyama2006active,
  title={Active learning in approximately linear regression based on conditional expectation of generalization error.},
  author={Sugiyama, Masashi and Ridgeway, Greg},
  journal={Journal of Machine Learning Research},
  volume={7},
  number={1},
  year={2006}
}

@article{goebel2025budgeted,
  title={Budgeted Online Active Learning with Expert Advice and Episodic Priors},
  author={Goebel, Kristen and Solow, William and Pesantez-Cabrera, Paola and Keller, Markus and Fern, Alan},
  journal={arXiv preprint arXiv:2506.03307},
  year={2025}
}

@article{howard2007information,
  title={Information value theory},
  author={Howard, Ronald A},
  journal={IEEE Transactions on systems science and cybernetics},
  volume={2},
  number={1},
  pages={22--26},
  year={2007},
  publisher={IEEE}

}

@article{cohn1994improving,
  title={Improving generalization with active learning},
  author={Cohn, David and Atlas, Les and Ladner, Richard},
  journal={Machine learning},
  volume={15},
  number={2},
  pages={201--221},
  year={1994},
  publisher={Springer}
}

@article{cohn1994active,
  title={Active learning with statistical models},
  author={Cohn, David and Ghahramani, Zoubin and Jordan, Michael},
  journal={Advances in neural information processing systems},
  volume={7},
  year={1994}
}

@article{cacciarelli2022stream,
  title={Stream-based active learning with linear models},
  author={Cacciarelli, Davide and Kulahci, Murat and Tyssedal, John S{\o}lve},
  journal={Knowledge-Based Systems},
  volume={254},
  pages={109664},
  year={2022},
  publisher={Elsevier}
}

@inproceedings{saran2023streaming,
  title={Streaming active learning with deep neural networks},
  author={Saran, Akanksha and Yousefi, Safoora and Krishnamurthy, Akshay and Langford, John and Ash, Jordan T},
  booktitle={International Conference on Machine Learning},
  pages={30005--30021},
  year={2023},
  organization={PMLR}
}

@article{hazan2007logarithmic,
  title={Logarithmic regret algorithms for online convex optimization},
  author={Hazan, Elad and Agarwal, Amit and Kale, Satyen},
  journal={Machine Learning},
  volume={69},
  number={2},
  pages={169--192},
  year={2007},
  publisher={Springer}
}

@article{hoi2021online,
  title={Online learning: A comprehensive survey},
  author={Hoi, Steven CH and Sahoo, Doyen and Lu, Jing and Zhao, Peilin},
  journal={Neurocomputing},
  volume={459},
  pages={249--289},
  year={2021},
  publisher={Elsevier}
}

@techreport{settles.tr09,
Author = {Burr Settles},
Institution = {University of Wisconsin--Madison},
Number = {1648},
Title = {Active Learning Literature Survey},
Type = {Computer Sciences Technical Report},
Year = {2009}
}

@article{pinson2022regression,
  title={Regression markets and application to energy forecasting},
  author={Pinson, Pierre and Han, Liyang and Kazempour, Jalal},
  journal={{TOP}},
  volume={30},
  number={3},
  pages={533--573},
  year={2022},
  publisher={Springer}
}

@article{dasgupta2007general,
  title={A general agnostic active learning algorithm},
  author={Dasgupta, Sanjoy and Hsu, Daniel J and Monteleoni, Claire},
  journal={Advances in neural information processing systems},
  volume={20},
  year={2007}
}

@article{cacciarelli2024active,
  title={Active learning for data streams: a survey},
  author={Cacciarelli, Davide and Kulahci, Murat},
  journal={Machine Learning},
  volume={113},
  number={1},
  pages={185--239},
  year={2024},
  publisher={Springer}
}

@article{li2024survey,
  title={A survey on deep active learning: Recent advances and new frontiers},
  author={Li, Dongyuan and Wang, Zhen and Chen, Yankai and Jiang, Renhe and Ding, Weiping and Okumura, Manabu},
  journal={IEEE Transactions on Neural Networks and Learning Systems},
  volume={36},
  number={4},
  pages={5879--5899},
  year={2024},
  publisher={IEEE}
}

@article{sabato2014active,
  title={Active regression by stratification},
  author={Sabato, Sivan and Munos, Remi},
  journal={Advances in Neural Information Processing Systems},
  volume={27},
  year={2014}
}

@inproceedings{hao2024composite,
  title={Composite active learning: Towards multi-domain active learning with theoretical guarantees},
  author={Hao, Guang-Yuan and Huang, Hengguan and Wang, Haotian and Gao, Jie and Wang, Hao},
  booktitle={Proceedings of the AAAI Conference on Artificial Intelligence},
  volume={38},
  number={11},
  pages={12286--12294},
  year={2024}
}

@article{luth2023navigating,
  title={Navigating the pitfalls of active learning evaluation: A systematic framework for meaningful performance assessment},
  author={L{\"u}th, Carsten and Bungert, Till and Klein, Lukas and Jaeger, Paul},
  journal={Advances in Neural Information Processing Systems},
  volume={36},
  pages={9789--9836},
  year={2023}
}

@article{kontonis2024active,
  title={Active classification with few queries under misspecification},
  author={Kontonis, Vasilis and Ma, Mingchen and Tzamos, Christos},
  journal={Advances in Neural Information Processing Systems},
  volume={37},
  pages={51684--51707},
  year={2024}
}

@article{fujii2016budgeted,
  title={Budgeted stream-based active learning via adaptive submodular maximization},
  author={Fujii, Kaito and Kashima, Hisashi},
  journal={Advances in Neural Information Processing Systems},
  volume={29},
  year={2016}
}

@inproceedings{agarwal2019marketplace,
  title={A marketplace for data: An algorithmic solution},
  author={Agarwal, Anish and Dahleh, Munther and Sarkar, Tuhin},
  booktitle={Proceedings of the 2019 ACM Conference on Economics and Computation},
  pages={701--726},
  year={2019}
}

@inproceedings{ghosh2011selling,
  title={Selling privacy at auction},
  author={Ghosh, Arpita and Roth, Aaron},
  booktitle={Proceedings of the 12th ACM conference on Electronic commerce},
  pages={199--208},
  year={2011}
}

@inproceedings{ghorbani2019data,
  title={Data shapley: Equitable valuation of data for machine learning},
  author={Ghorbani, Amirata and Zou, James},
  booktitle={International conference on machine learning},
  pages={2242--2251},
  year={2019},
  organization={PMLR}
}

@inproceedings{yoon2020data,
  title={Data valuation using reinforcement learning},
  author={Yoon, Jinsung and Arik, Sercan and Pfister, Tomas},
  booktitle={International Conference on Machine Learning},
  pages={10842--10851},
  year={2020},
  organization={PMLR}
}

@inproceedings{zinkevich2003online,
  title={Online convex programming and generalized infinitesimal gradient ascent},
  author={Zinkevich, Martin},
  booktitle={International Conference on Machine Learning (ICML)},
  pages={928--936},
  year={2003}
}

@article{shalev2012online,
  title={Online learning and online convex optimization},
  author={Shalev-Shwartz, Shai},
  journal={Foundations and Trends® in Machine Learning},
  volume={4},
  number={2},
  pages={107--194},
  year={2012},
  publisher={Now Publishers, Inc.}
}

@inproceedings{wimalaratne2022unisolar,
  title={UNISOLAR: An Open Dataset of Photovoltaic Solar Energy Generation in a Large Multi-Campus University Setting},
  author={Wimalaratne, S. and Haputhanthri, D. and Kahawala, S. and Gamage, G. and Alahakoon, D. and Jennings, A.},
  booktitle={2022 15th International Conference on Human System Interaction (HSI)},
  pages={1--5},
  year={2022},
  organization={IEEE}
}

@article{huang2026purchase,
  title={How to Purchase Labels? A Cost-Effective Approach Using Active Learning Markets},
  author={Huang, Xiwen and Pinson, Pierre},
  journal={INFORMS Journal on Data Science},
  year={2026},
  publisher={INFORMS}
}

@techreport{settles2009survey,
  title       = {Active Learning Literature Survey},
  author      = {Settles, Burr},
  institution = {University of Wisconsin--Madison},
  year        = {2009}
}

@article{balcan2010true,
  title={The true sample complexity of active learning},
  author={Balcan, Maria-Florina and Hanneke, Steve and Vaughan, Jennifer Wortman},
  journal={Machine learning},
  volume={80},
  number={2},
  pages={111--139},
  year={2010},
  publisher={Springer}
}

@inproceedings{osogami2021second,
  title={Second order techniques for learning time-series with structural breaks},
  author={Osogami, Takayuki},
  booktitle={Proceedings of the AAAI Conference on Artificial Intelligence},
  volume={35},
  number={10},
  pages={9259--9267},
  year={2021}
}

@article{luo2016efficient,
  title={Efficient second order online learning by sketching},
  author={Luo, Haipeng and Agarwal, Alekh and Cesa-Bianchi, Nicolo and Langford, John},
  journal={Advances in Neural Information Processing Systems},
  volume={29},
  year={2016}
}

@misc{orabona2025modernintroductiononlinelearning,
      title={A Modern Introduction to Online Learning}, 
      author={Francesco Orabona},
      year={2025},
      eprint={1912.13213},
      archivePrefix={arXiv},
      primaryClass={cs.LG},
      url={https://arxiv.org/abs/1912.13213}, 
}

@article{pandey2023strategic,
  title={Strategic coalition for data pricing in IoT data markets},
  author={Pandey, Shashi Raj and Pinson, Pierre and Popovski, Petar},
  journal={IEEE Internet of Things Journal},
  volume={11},
  number={4},
  pages={6454--6468},
  year={2023},
  publisher={IEEE}
}

% \bibitem{ref1}
% {\it{Mathematics Into Type}}. American Mathematical Society. [Online]. Available: https://www.ams.org/arc/styleguide/mit-2.pdf

% \bibitem{ref2}
% T. W. Chaundy, P. R. Barrett and C. Batey, {\it{The Printing of Mathematics}}. London, U.K., Oxford Univ. Press, 1954.

% \bibitem{ref3}
% F. Mittelbach and M. Goossens, {\it{The \LaTeX Companion}}, 2nd ed. Boston, MA, USA: Pearson, 2004.

% \bibitem{ref4}
% G. Gr\"atzer, {\it{More Math Into LaTeX}}, New York, NY, USA: Springer, 2007.

% \bibitem{ref5}M. Letourneau and J. W. Sharp, {\it{AMS-StyleGuide-online.pdf,}} American Mathematical Society, Providence, RI, USA, [Online]. Available: http://www.ams.org/arc/styleguide/index.html

% \bibitem{ref6}
% H. Sira-Ramirez, ``On the sliding mode control of nonlinear systems,'' \textit{Syst. Control Lett.}, vol. 19, pp. 303--312, 1992.

% \bibitem{ref7}
% A. Levant, ``Exact differentiation of signals with unbounded higher derivatives,''  in \textit{Proc. 45th IEEE Conf. Decis.
% Control}, San Diego, CA, USA, 2006, pp. 5585--5590. DOI: 10.1109/CDC.2006.377165.

% \bibitem{ref8}
% M. Fliess, C. Join, and H. Sira-Ramirez, ``Non-linear estimation is easy,'' \textit{Int. J. Model., Ident. Control}, vol. 4, no. 1, pp. 12--27, 2008.

% \bibitem{ref9}
% R. Ortega, A. Astolfi, G. Bastin, and H. Rodriguez, ``Stabilization of food-chain systems using a port-controlled Hamiltonian description,'' in \textit{Proc. Amer. Control Conf.}, Chicago, IL, USA,
% 2000, pp. 2245--2249.

% \end{thebibliography}

\newpage

\vfill

\end{document}